# Pragmatics beyond humans:
# meaning, communication, and LLMs

Vít Gvoždiak[1]

**Abstract**

The paper reconceptualizes pragmatics not as a subordinate, third dimension of meaning, but as a dynamic interface through which language operates as a socially embedded tool for action. With the emergence of large language models (LLMs) in communicative contexts, this understanding needs to be further refined and methodologically reconsidered. The first section challenges the traditional semiotic trichotomy, arguing that connectionist LLM architectures destabilize established hierarchies of meaning, and proposes the Human-Machine Communication (HMC) framework as a more suitable alternative. The second section examines the tension between human-centred pragmatic theories and the machine-centred nature of LLMs. While traditional, Gricean-inspired pragmatics continue to dominate, it relies on human-specific assumptions ill-suited to predictive systems like LLMs. Probabilistic pragmatics, particularly the Rational Speech Act framework, offers a more compatible teleology by focusing on optimization rather than truth-evaluation. The third section addresses the issue of substitutionalism in three forms – generalizing, linguistic, and communicative – highlighting the anthropomorphic biases that distort LLM evaluation and obscure the role of human communicative subjects. Finally, the paper introduces the concept of "context frustration" to describe the paradox of increased contextual input paired with a collapse in contextual understanding, emphasizing how users are compelled to co-construct pragmatic conditions both for the model and themselves. These arguments suggest that pragmatic theory may need to be adjusted or expanded to better account for communication involving generative AI.

## 0 Introduction

Linguistic communication does not consist solely in the production and interpretation of sequences of words according to grammatical rules, but trivially it also involves a range of linguistically communicative aspects that cannot be captured exclusively through lexico-syntactic principles. With the advent and popularity of large language models (LLMs) and their chat forms (especially ChatGPT released in 2022), new questions have emerged regarding whether and/or to what extent LLMs are capable of handling communicative phenomena whose nature goes beyond lexico-semantic de-coding and relies on contextual and pragmatic processes. For pragmatics and pragmatic inquiry, as for many other disciplines, current technological developments present both a challenge and an opportunity for methodological extension and possible evaluation of its assumptions, practices and aims.

Pragmatics itself is a significantly multidisciplinary field, shaped by contributions from linguistics, philosophy, communication theory, cognitive science, psychology, sociology, &c.[2] Its ambiguous boundaries often lead to the perception of pragmatics as a kind of "wastebasket" of syntax or semantics. According to Gazdar's famous definition, pragmatics deals with "those aspects of the meaning of utterances which cannot be accounted for by straightforward reference to the truth conditions of the sentences uttered. Put crudely: PRAGMATICS = MEANING - TRUTH CONDITIONS"[3]. Over the fifty years since its formulation and even before the emergence of LLMs, this definitional equation had undergone many revisions. These revisions include re-thinking the parameters involved in understanding (e.g. by reinforcing it or challenging it) Gazdar's "minus" in truth conditions, i.e. the principles of distinguishing in particular between semantic and pragmatic mechanisms[4], moving from purely theoretical or armchair approaches to experimental methods[5], and last but not least examining the assumptions and consequences of non-ideal aspects of communication[6]. And unsurprisingly, with the advent of LLMs, the theoretical beliefs and specific experimental practices of pragmatics continue to change significantly.[7]

---

[1] Institute of Philosophy, Czech Academy of Sciences, Prague. Email: gvozdiak@flu.cas.cz. ORCID: https://orcid.org/0000-0002-0484-8127
[2] Cf. e.g. Anat Biletzki, "Is there a history of pragmatics?" *Journal of Pragmatics* 25, no. 4 (1996): 455-470; Wataru Koyama, "The rise of pragmatics: a historiographic overview," in *Foundations of Pragmatics*, eds. Wolfram Bublitz and Neal R. Norrick (Berlin – Boston: De Gruyter Mouton, 2011), 139–166.

This paper addresses four broad but interrelated issues that I argue characterize these ongoing transformations of/in pragmatics. In the first section, against the backdrop of the traditional semiotic trichotomy, I address the question of the role of pragmatics in assessing the meaningfulness of LLMs' linguistic communicative performances and argue that the qualitative nature of these performances offer a valuable opportunity to move beyond the discrete semiotic trichotomy in favour of the Human–Machine Communication (HMC) framework which provides a more methodologically relevant approach. On this basis, the following sections explore three types of tensions or paradoxes that arise in pragmatic research on LLMs. In the second section, I attempt to point to the fact that current research on LLMs' pragmatic abilities often relies on human-centred pragmatic theories that overlook the distinctive nature of LLMs and to highlight the benefits of an alternative framework of probabilistic pragmatics. The third section addresses the discursive and methodological problem of substitutionalism, i.e. the mechanism of substituting different aspects of communicative processes (but primarily human and LLM subjects) which creates a significant methodological asymmetry in favour of the role of machines, while diminishing attention to changes in human linguistic and pragmatic behaviour. And finally in the fourth section, I discuss the tension between different conceptions of context and examine their implications for pragmatics in the form of what I call context frustration.

## 1 Pragmatics: semiotic trichotomy and HMC

The place of pragmatics and pragmatic aspects of utterances has traditionally[8] been situated within the semiotic prism of meaning constitution, where the meaning of a sign (word, phrase, sentence/utterance) is understood as a result of a complex relational structure in which (i) a syntactic relation to other signs, (ii) a semantic relation to what the sign represents, and (iii) a pragmatic relation to the user of the sign are defined. These levels contribute to the constitution of meaning in a hierarchical fashion, i.e. as David Kaplan, following Rudolf Carnap, argues: "the overall theory of language should be constructed with syntax at the base, semantics built upon that, and pragmatics built upon semantics"[9]. At the same time, however, the meaning and meaningfulness of a sign cannot be unambiguously located within any single dimension. Any attempt at a locational reduction can at best offer a description of certain mechanisms of signification, but not to an exhaustive description of signification in its entirety and *en tout*.

---

[3] Gerald Gazdar, *Pragmatics* (New York – London: Academic Press, 1979), 2.
[4] There is a fairly rich literature on this issue and, for simplicity, we can define two camps, but it is true that these camps are not homogeneous and members of one camp may differ in a number of ways. On the one hand, there are those who are convinced of the existence of minimal truth-conditional content as a consequence of lexico-syntactic rules without the more fundamental influence of context. Cf. e.g. Emma Borg, *Minimal Semantics* (Oxford: Oxford University Press, 2004); Emma Borg, *Pursuing Meaning* (Oxford: Oxford University Press, 2012); Herman Cappelen and Ernie Lepore, *Insensitive Semantics* (Malden – Oxford – Carlton: Blackwell, 2005); Michael Devitt, *Overlooking Conventions* (Cham: Springer, 2021). On the other hand, there are those who are convinced that no propositional content exists without the contribution of contextual elements. Cf. e.g. Francois Recanati, *Literal Meaning* (New York: Cambridge University Press, 2004); Francois Recanati, *Truth-Conditional Pragmatics* (Oxford: Clarendon Press, 2010); Robyn Carston, *Thoughts and Utterances* (Oxford – Berlin: Blackwell, 2002); Dan Sperber and Deirdre Wilson, *Relevance. Communication and Cognition* (Oxford – Cambridge: Blackwell, 1995).
[5] Cf. e.g. Ira A. Noveck and Dan Sperber, eds., *Experimental Pragmatics* (Basingstoke – New York: Palgrave, 2004); Ira A. Noveck and Anne Reboul, "Experimental Pragmatics: a Gricean turn in the study of language," *Trends in Cognitive Science* 12, no. 11 (2008): 425–431.
[6] Cf. e.g. Herman Cappelen and Josh Dever, *Bad Language* (Oxford: Oxford University Press, 2019); Jessica Keiser, *Non-Ideal Foundations of Language* (New York: Routledge, 2023).
[7] For overview of topics and methods cf. e.g. Bolei Ma et al., "Pragmatics in the Era of Large Language Models: A Survey on Datasets, Evaluation, Opportunities and Challenges". In *Proceedings of the 63rd Annual Meeting of the Association for Computational Linguistics (Volume 1: Long Papers)*, eds. Wanxiang Che et al. (Vienna: Association for Computational Linguistics, 2025), 8679–8696.
[8] Charles W. Morris, *Foundations of the Theory of Signs* (Chicago: University of Chicago Press, 1938); Rudolf Carnap, *Introduction to Semantics* (Cambridge: Harvard University Press, 1948).

This semiotic trichotomy, in the context of AI and LLMs, often functions as a general framework through which it is easy to formulate basic forms of arguments that, at a theoretical level, aim to show, that LLMs (and other non-human systems) cannot be regarded as entities capable of understanding linguistic utterances and producing fully meaning-saturated utterances, since they are unable to engage with some (or even all) of these dimensions.

Most notably, these debates tend to cluster at the border between syntax and semantics, where we can find both Searle's[10] famous Chinese Room thought experiment / argument and its more recent reformulations, such as the Octopus Test[11] and the National Library of Thailand[12] thought experiments. All of these attempt to show, in one sense or another, that computers (or LLMs) operate solely on the level of syntactic relations, and given the hierarchical nature of the semiotic trichotomy, it can be easily argued that rejecting LLMs' semantic competence necessarily entails the rejection of their pragmatic competence as well.

Even more radically, the Chomskyan stance within the semiotic trichotomy questions whether we can meaningfully speak of adequate syntactic abilities in the case of LLMs at all. Andrea Moro et al.[13] attempt to show that LLMs are not only incapable of analyzing the underlying syntactic structures of linguistic utterances, but due to the nature of their learning processes and input-output operations, there is, unlike for human speakers, no distinction between possible and impossible languages, i.e. between those based on the hierarchical and recursive grammars of natural languages and those that follow merely linear rules, which are not found in any natural language. From this perspective, one could say that LLMs do not meaningfully engage with language because they lack even the foundational level of syntax.[14]

If we consistently follow the logic of this hierarchical trichotomy, we might claim that the Turing[15] imitation game is not only a conceptual-behavioral test, but in fact a test of global understanding, and therefore essentially pragmatic since pragmatics is the ultimate culmination of the semiotic hierarchy. From this viewpoint, Searle's anti-functionalist rejection of the semantic capacities of computers can be read *de facto* as grounded in a Morrisian concept of pragmatics, when he states that, unlike the computer, human is "able to understand English [...] because [...] [he is] a certain sort of organism with a certain biological (i.e. chemical and physical) structure"[16].

Pushing this further, one could argue that syntactic-semantic objections, such as those raised by the aforementioned thought experiments or the symbol grounding problem[17] (SGP), cannot be considered a real problem not only with regard to human speakers but also and more importantly in the case of LLMs. According to Reto Gubelmann[18], we can replace SGP and its semantic assumption that the meaningfulness of utterances is grounded in their relation to other (extra-linguistic) entities with a framework based on pragmatic norms of linguistic behaviour and the assumption that utterance meanings are grounded in the community practices of speakers. In this light words, both positive and negative arguments can be broadly described as pragmatic, in Searle's case as a rejection based on the notion that LLMs do not constitute entities capable of entering into meaningful pragmatic relations; in Gubelmann's case, as a methodological commitment to the view that pragmatic norms, as fundamental principles of meaningfulness of utterances, leave traces in the training data that LLMs are in principle capable of learning and processing.

---

[9] David Kaplan, "Afterthoughts," in *Themes from Kaplan,* eds. Joseph Almog, John Perry and Howard Wettstein (New York – Oxford: Oxford University Press, 1989), 576.

[10] John Searle, "Minds, Brains and Programs," *Behavioral and Brain Sciences* 3, no. 3 (1980): 417–457.

[11] Emily Bender and Alexander Koller, "Climbing towards NLU: On Meaning, Form, and Understanding in the Age of Data," in *Proceedings of the 58th Annual Meeting of the Association for Computational Linguistics*, eds. Dan Jurafsky et al. (Online. Association for Computational Linguistics, 2020), 5185–5198.

[12] Emily Bender, "Thought experiment in the National Library of Thailand," *Medium*, accessed March 15, 2025, https://medium.com/@emilymenonbender/thought-experiment-in-the-national-library-of-thailand-f2bf761a8a83.

[13] Andrea Moro, Matteo Greco and Stefano F. Cappa, "Large languages, impossible languages and human brains," *Cortex* 167 (2023): 82–85.

[14] However, based on indications from empirical findings, training LLMs (or a relatively small model GPT-2) on impossible languages seems to lead to significantly worse results compared to the possible/actual language (English). Cf. Julie Kallini et al., "Mission: Impossible Language Models," in *Proceedings of the 62nd Annual Meeting of the Association for Computational Linguistics (Volume 1: Long Papers)*, eds. Lun-Wei Ku, Andre Martins and Vivek Srikumar (Bangkok: Association for Computational Linguistics, 2024), 14691–14714.

[15] Alan Turing, "Computing Machinery and Intelligence," *Mind* 59, no. 236 (1950): 433–460.

[16] John Searle, "Minds, Brains and Programs," *Behavioral and Brain Sciences* 3, no. 3 (1980): 422.

[17] Steven Harnad, "The Symbol Grounding Problem," *Physica* D 42 (1990): 335–346. For LLMs specifically cf. e.g. Dimitri Coelho Mollo and Raphaël Millière, "The Vector Grounding Problem," arXiv:2304.01481 [cs.CL], 2023.

[18] Reto Gubelmann, "Pragmatic Norms Are All You Need – Why The Symbol Grounding Problem Does Not Apply to LLMs," in *Proceedings of the 2024 Conference on Empirical Methods in Natural Language Processing* (Miami: Association for Computational Linguistics, 2024), 11663–11678.

One of the consequences of the advent of LLMs for pragmatics is therefore the relativization of the Carnapian/Kaplanian belief in the strict hierarchy and clear-cut boundaries of the semiotic trichotomy, and the conceptual revision of the syntactic-semantic-pragmatic framework along with the recognition that the its formal elegance cannot be projected into the description of communicative processes as a discrete dimensions of meaningfulness.[19]

Alongside the established relationship between utterance compositionality (its syntactic-semantic structure and the overall hierarchical view of the semiotic trichotomy), there persists a largely unexamined assumption that the semiotic trichotomy is viewed through a symbolic lens, one rooted in explicit rules and discrete combinatorial units. This view fails to account for the connectionist nature that underlies modern AI/LLMs technologies. What are the consequences of the fact that language and its use may not be based on discrete symbolic units and fixed combinatorial rules? And what does this mean for the traditional distinction between mechanisms responsible for what is said by convention and those responsible for what is communicated by other means?

From the standpoint of pragmatics as the dimension that arguably completes the architecture of meaning-making process, it is perhaps unsurprising that behavioural tests of understanding, such as Turing's imitation game, are increasingly seen as inadequate. This inadequacy is supported by recent empirical findings. On the one hand, human participants often struggle to reliably determine whether a speaker is human or an AI[20], even when explicitly pragmatic criteria are applied[21]. On the other hand, however, in more tightly controlled experimental contexts (such as debate-based consensus games)[22], the linguistic and communicative behaviour of LLMs still reveals significant differences when compared to that of humans.

It is essential to stress from the outset that the pragmatic level, unlike syntax and semantics, relies on a form of implicit or explicit communicative situatedness and conversational context, as all the thought experiments mentioned above demonstrate. Thus pragmatics is not only connected to the syntactic and semantic dimensions (in the form that we can call linguistic pragmatics or pragmalinguistics), but also to psychology and cognitive sciences[23], and more broadly to communication parameters[24], which manifest concretely in the organisation of communicative practices (in the sociopragmatic sense) and in the various forms in which LLMs may participate in these practices. This aspect is particularly captured by human-machine communication (HMC) approaches, which assume that communication is not exclusively a human activity and that machines and technologies are not only communicative tools, but may also act as active communicative entities that contribute to the meaning-making processes.

---

[19] In these types of accounts, we find a whole spectrum of positions concerning the illusion of descriptive adequacy of syntactic-semantic dichotomy, often emphasizing alternative interpretations based on inferential semantics and inferential role semantics. Cf. e.g. Jaroslav Peregrin, "Do Computers "Have Syntax, But No Semantics"?" *Minds and Machines* 31, no. 2 (2021): 305–321; Steven T. Piantadosi and Felix Hill, "Meaning without reference in large language models," arXiv:2208.02957 [cs.CL], 2022; Vladimír Havlík, "Meaning and understanding in large language models," *Synthese* 205, no. 9 (2025).

[20] Kristina Radivojevic, Nicholas Clark and Paul Brenner, "LLMs Among Us: Generative AI Participating in Digital Discourse," arXiv:2402.07940 [cs.HC], 2024.

[21] Xi Chen, Jun Li and Yuting Ye, "A feasibility study for the application of AI-generated conversations in pragmatic analysis," *Journal of Pragmatics* 223 (2024): 14-30.

[22] James Flamino et al., "Testing the limits of large language models in debating humans," *Scientific Reports* 15, 13852 (2025).

[23] Cf. e.g. Bram van Dijk et al., "Large Language Models: The Need for Nuance in Current Debates and a Pragmatic Perspective on Understanding," in P*roceedings of the 2023 Conference on Empirical Methods in Natural Language Processing*, eds. Houda Bouamor, Juan Pino, Kalika Bali (Singapore: Association for Computational Linguistics, 2023), 12641–12654.

[24] Cf. e.g. David J. Gunkel, *AI for Communication* (Boca Raton: CRC Press, 2025).

For the pragmatics of communication, which explicitly reflects the implications of AI for conversational practices, three main research strands emerge, according to Andrea Guzman and Seth Lewis[25]: (i) functional, (2) relational, and (3) metaphysical.

The functional aspect adresses with the fundamental principles on the basis of which AI can become (or in fact becomes) a conversational partners. It accounts for different types of communicative genres, e.g. the distinciton between interpersonal and mass communication, and seeks to critically redefine these in light of the fact that existing human communicative practices may not provide a fully adequate analogy for the conversations in which AI participates. A central methodological concern here is the overly mechanistic nature of human linguistic behaviour which is often used as the gold standard in evaluating LLMs performance. According to Andrea Guzman and Seth Lewis, the communicative position of AI can be described as hybrid, i.e., users recognise AI as a machine, but often perceive it through the human-like attributes, particularly voice and gendered linguistic cues. Defining the specific characteristics of AI as a communicating entity forms the basis of functionally oriented research aimed at mapping other generally pragmatic features of communication, such as the context-sensitive ways in which e.g. time and space are expressed.

The relational aspect concerns the influence of AI technologies on how human participants perceive and negotiate their own communicative roles, especially given that the historically and culturally constructed roles of humans often serve as a blueprint for assigning roles to AI. Andrea Guzman and Seth Lewis argue that assuming such originally human-specific roles entails not only the need to explore different types of power relations between humans and technology, but also has direct consequences for the identity constitution of human conversational participants, potentially leading to the degradation of human roles and to the disruption of (some) social, cultural and political processes. In this light, it is not only the practices and design choices of particular technological solutions that matter, but also the ways in which AI is debated in public, since public debates become a space where relational aspects of AI are negotiated.

The metaphysical aspect studies the implications of the entry of AI into conversational exchanges for the traditional distinction between humans as communicating subjects and technologies as communicative tools. This distinction has become significantly blurred in the context of the rapid development of AI and LLMs in recent years, further reinforcing the nature of reflections on the nature of technology and humans. This has also brought ethical and legislative issues into focus, particularly the inadequacy of existing anthropomorphic framework. More importantly it reveals that the most fundamental change may be not only in what we think of "technology" and "human", but in acknowledging that communication itself is no longer an exclusively human process and that communication performances now also increasingly involve machines.

Rather than defining pragmatics by its position and role within the semiotic trichotomy, I find it more productive to frame its tasks through the threefold agenda proposed by this HMC framework. The functional, relational, and metaphysical aspects (as it is clearly evident from the preceding discussion) often overlap in significant ways and cannot be straightforwardly mapped onto the traditional semiotic dimensions. Nevertheless, in my view they offer a more suitable starting point for discussing the current forms of pragmatics, especially those related to LLMs. In the following three sections, I draw loosely on this tripartite structure and translate it into a more concrete articulation of three interconnected issues: (1) the anthropocentric orientation of many pragmatic approaches, which corresponds primarily to the functional aspect; (2) the machine-centred focus of both theoretically and empirically driven LLM research, which connects particularly with the relational and metaphysical aspects; and (3) the misalignment in the understanding and operation of human and machine contexts, which concerns mainly the functional and metaphysical dimensions.

---

[25] Andrea L. Guzman and Seth C. Lewis, "Artificial intelligence and communication: A Human–Machine Communication research agenda," *New Media & Society* 22, no. 1 (2019): 70-86.

## 2 Human-centred pragmatics

Most pragmatic approaches to LLMs can be characterised as an extrapolation or specification of what Marco Baroni[26] calls linguistically-oriented deep net analysis. Within the pragmatic framework, this primarily means empirically testing selected problems that stem from the conceptual-theoretical background of traditional pragmatic theories in order to asses model behaviour within a chosen domain, typically through comparison with human speakers. The domains investigated in this framework, and therefore treated as pragmatic, can be classified in various ways.[27] In general, oe may follow Robert Stalnaker's[28] view that pragmatics is primarily concerned with speech acts and the contexts in which they are performed and define two principal types of pragmatic problems. The first concerns the necessary and sufficient conditions for performing a speech act, which are distinct from the truth conditions of the proposition expressed. The second problem concerns the influence of the linguistic context on the proposition that the speaker expresses by using the sentence in that context.

Stalnaker's first problem in current pragmatic research on LLMs corresponds to (i) the investigation of the interpretative-performative mechanisms by which the speaker performs a certain act through the production of words, sentences and texts as a result of socio-cultural conditions and is identified as such by the addressee. When applied to LLMs, more general discussions tend to focus on arguments why LLMs, despite producing valid linguistic outputs, cannot genuinely perform speech acts[29], or more narrowly, why they may only be capable of performing some of them (namely, assertions)[30]. Empirical research within this framework centres, among other things, on the model's ability to recognise different types of speech acts, e.g. apologies[31], or in its sensitivity to expressions of hate speech[32] and impoliteness[33].

The second Stalnakerian problem can be further subdivided into distinct types of pragmatic mechanisms: (ii) pre-propositional mechanisms, which are an essential part of the process by which propositional content is constituted, and which concern expressions requiring contextual saturation, a process that is typically triggered and constrained by linguistic cues; in LLMs research, this usually involves various forms of ambiguity[34]; (iii) post-propositional mechanisms, which enable a speaker to convey, through the use of a sentence, not only its linguistically compositional literal proposition, but also non-literal contents, most commonly various types of implicatures[35]; and (iv) inter-propositional mechanisms, which contribute to the cohesion and coherence of larger textual structures, for example, by organising inter-utterance relations or track changes in entity states throughout a discourse[36].

---

[26] Marco Baroni, "On the proper role of linguistically-oriented deep net analysis in linguistic theorizing," in *Algebraic systems and the representation of linguistic knowledge*, ed. Shalom Lappin (Abingdon-on-Thames: Taylor and Francis, 2022), 5-22.

[27] For example, Bolei Ma et al. divide pragmatic research of LLMs into five main areas: (i) context and deixis, (ii) implicature and presupposition, (iii) speech acts and intention recognition, (iv) discourse and coherence, and (v) social pragmatics. Cf. Bolei Ma et al., "Pragmatics in the Era of Large Language Models: A Survey on Datasets, Evaluation, Opportunities and Challenges". In *Proceedings of the 63rd Annual Meeting of the Association for Computational Linguistics (Volume 1: Long Papers)*, eds. Wanxiang Che et al. (Vienna: Association for Computational Linguistics, 2025), 8681–8682. Settaluri Sravanthi et al. divide the field of pragmatics into (i) implicatures, (ii) presuppositions, (iii) reference, and (iv) deixis. Cf. Settaluri Sravanthi et al., "PUB: A Pragmatics Understanding Benchmark for Assessing LLMs' Pragmatics Capabilities," in *Findings of the Association for Computational Linguistics: ACL 2024*, eds. Lun-Wei Ku, Andre Martins and Vivek Srikumar (Bangkok: Association for Computational Linguistics, 2024), 12075–12097. Other classifications are more specific, for example, Rui Ma et al. define five categories of pragmatic processing as (1) metaphor understanding, (2) sarcasm detection, (3) personality recognition, (4) aspect extraction, and (5) polarity detection. Cf. Rui Mao et al., "A survey on pragmatic processing techniques," *Information Fusion* 114, 102712 (2025). All these classifications and differentiations of the field of pragmatic mechanisms are not guided by any explicit methodology and are probably based more on common knowledge with a strong gravitation towards a Gricean-type conception of cooperative nature.

[28] Robert C. Stalnaker, *Context and Content* (Oxford: Oxford University Press, 1999): 34.

[29] Cf. e.g. Reto Gubelmann, "Large Language Models, Agency, and Why Speech Acts are Beyond Them (For Now) – A Kantian-Cum-Pragmatist Case," *Philosophy & Technology* 37, 32 (2024); Zachary P Rosen and Rick Dale, "LLMs Don't "Do Things with Words" but Their Lack of Illocution Can Inform the Study of Human Discourse," *Proceedings of the Annual Meeting of the Cognitive Science Society* 46 (2024).

[30] Iwan Williams and Tim Bayne, "Chatting with bots: AI, speech acts, and the edge of assertion," *Inquiry* (2024): 1–24.

[31] Danni Yu et al., "Assessing the potential of LLM-assisted annotation for corpus-based pragmatics and discourse analysis: The case of apology," *International Journal of Corpus Linguistics* 29, no. 4 (2024): 534–561.

[32] Min Zhang et al., "Don't Go To Extremes: Revealing the Excessive Sensitivity and Calibration Limitations of LLMs in Implicit Hate Speech Detection," in *Proceedings of the 62nd Annual Meeting of the Association for Computational Linguistics (Volume 1: Long Papers)*, eds. Lun-Wei Ku, Andre Martins, Vivek Srikumar (Bangkok: Association for Computational Linguistics, 2024), 12073–12086.

[33] Marta Andersson and Dan McIntyre, "Can ChatGPT recognize impoliteness? An exploratory study of the pragmatic awareness of a large language model," *Journal of Pragmatics* 239 (2025): 16-36.

[34] Cf. e.g. Sewon Min et al., "AmbigQA: Answering Ambiguous Open-domain Questions," in *Proceedings of the 2020 Conference on Empirical Methods in Natural Language Processing (EMNLP)*, eds. Bonnie Webber et al. (Online:

The theoretical basis for such research is typically formed by human-centred or anthropomorphic pragmatic theories. A paradigmatic example is the use of H. P. Grice's[37] theory of cooperative communication, which, especially through the lens of conversational maxims, assesses whether models can perform implicature inference. Empirical studies usually rest on the assumption that the generation and interpretation of implicatures relies on the specifically human rationality of the participants who observe the cooperative principle. However, the assumptions guiding human engagement in interaction may differ (or even *de facto* differ) from those according to which AI and LLMs operate.

If one accepts that "the norm of LLM outputs is word occurrence probability, not truth"[38], then any pragmatic theory relying on the notion of truth in one sense or another (as, for instance, the Gricean account of what is said in relation to implicatures) clearly introduces a methodological bias. Pragmatic tasks assigned to LLMs might therefore be more accurately described not as an assessment of how adequate the outputs are in terms of propositional truth and possible contextual adjustments and inferences based on the assumption that communicators adhere to a rational principle of cooperation, but rather as "how to perform next-word prediction on samples of Internet text, given the mechanisms available in a neural network"[39]. R. Thomas McCoy et al.[40] argue that the particular functioning of LLMs should be taken seriously, and that their capacities should not to be tested in the same way as human capabilities. Accordingly, they propose and substantiate through a number of experiments a general teleological approach grounded in three types of sensitivity: (i) sensitivity to the frequency of the task presented to LLMs – regardless of complexity, LLMs perform better on tasks that are frequent in the training data than on those that are less frequent, (ii) sensitivity to the probability of the output – even in deterministic tasks, LLMs perform better, if the probability of the output text is higher rather than lower, and (iii) sensitivity to the input probability – even in deterministic tasks, LLMs perform better in some cases if the probability of the input prompt is higher rather than lower, although this factor plays a less important role than (ii)).

Although this teleological approach constitutes a concrete methodological framework and a practical implementation of the functional aspect of LLMs and AI research (as discussed in Section 1), it remains

---

Association for Computational Linguistics, 2020), 5783–5797; Gaurav Kamath et al., "Scope Ambiguities in Large Language Models," *Transactions of the Association for Computational Linguistics* 12 (2024): 738–754; Pierpaolo Basile et al., "Exploring the Word Sense Disambiguation Capabilities of Large Language Models," arXiv:2503.08662 [cs.CL], 2025.

[35] Cf. e.g. Ljubiša Bojić, Predrag Kovačević and Milan Čabarkapa, "Does GPT-4 surpass human performance in linguistic pragmatics?" *Humanities and Social Sciences Communications* 12, 794 (2025); Yan Cong, "Manner implicatures in large language models," *Sci Rep* 14, 29113 (2024); Rashid Nizamani, Sebastian Schuster and Vera Demberg, "SIGA: A Naturalistic NLI Dataset of English Scalar Implicatures with Gradable Adjectives," in *Proceedings of the 2024 Joint International Conference on Computational Linguistics, Language Resources and Evaluation (LREC-COLING 2024)*, eds. Nicoletta Calzolari et al. (Torino: ELRA and ICC, 2024), 14784–14795; Laura Ruis et al., "The Goldilocks of Pragmatic Understanding: Fine-Tuning Strategy Matters for Implicature Resolution by LLMs," arXiv:2210.14986 [cs.CL], 2023; Yue Shisen et al., "Do Large Language Models Understand Conversational Implicature- A case study with a Chinese sitcom," in *Proceedings of the 23rd Chinese National Conference on Computational Linguistics (Volume 1: Main Conference)*, eds. Maosong Sun et al. (Taiyuan: Chinese Information Processing Society of China, 2024), 1270–1285; Zhuang Qiu, Xufeng Duan and Zhenguang Cai, "Does ChatGPT Resemble Humans in Processing Implicatures?" in *Proceedings of the 4th Natural Logic Meets Machine Learning Workshop (NALOMA23)*, eds. Stergios Chatzikyriakidis and Valeria de Paiva (Nancy: Association for Computational Linguistics, 2023), 25–34.

[36] Najoung Kim and Sebastian Schuster, "Entity Tracking in Language Models," in *Proceedings of the 61st Annual Meeting of the Association for Computational Linguistics (Volume 1: Long Papers)*, eds. Anna Rogers, Jordan Boyd-Graber and Naoaki Okazaki (Toronto: Association for Computational Linguistics, 2023): 3835–3855; Polina Tsvilodub et al, "Experimental Pragmatics with Machines: Testing LLM Predictions for the Inferences of Plain and Embedded Disjunctions," arXiv:2405.05776 [cs.CL], 2024.

[37] Paul Grice, *Studies in the Way of Words* (Cambridge – London: Harvard University Press, 1989).

[38] Emma Borg, "LLMs, Turing tests and Chinese rooms: the prospects for meaning in large language models," *Inquiry* (2025): 24.

[39] R. Thomas McCoy et al., "Embers of Autoregression: Understanding Large Language Models Through the Problem They are Trained to Solve," arXiv:2309.13638 [cs.CL], 2023, 6.

[40] R. Thomas McCoy et al., "Embers of Autoregression: Understanding Large Language Models Through the Problem They are Trained to Solve," arXiv:2309.13638 [cs.CL], 2023.

largely unreflected in empirical pragmatic research. Among existing approaches, probabilistic pragmatics most closely approximates this functional-teleological orientation.

Probabilistic pragmatics is based on the assumption that language use is shaped by a range of contextual factors, whose defining feature is uncertainty. Based on this assumption, Michael Franke and Gerhard Jäger[41] define their framework for pragmatic theory as (i) probabilistic, (ii) interactive, (iii) rationalistic, (iv) computational, and (v) data-oriented. In other words: Uncertainties inherent in communicative exchanges can be modeled using (i) probability distributions as formal representations of linguistic expectations and decision making. Such exchanges are fundamentally (ii) relational and interactive. Probabilistic pragmatics explicitly considers the distinct production and interpretation perspectives of speakers and hearers in relation to context. It further treats (linguistic) behaviour as grounded in (iii) rational or optimal responses to communicative goals, while acknowledging that optimality is only ever approximate and constrained by cognitive limitations of language users. Its aim is to offer a mathematical model that is predictive, interpretable, and empirically testable. In this sense, its key feature is (iv) computability, which must necessarily be grounded in (v) empirical data that the model is designed to explain. Overall, the core objective of this approach is to frame pragmatic phenomena as non-discrete, continuous processes and to demonstrate that language users can learn such mechanisms from data.

A more concrete instantiation of probabilistic pragmatics is the Rational Speech Act[42] (RSA) framework, which is grounded primarily in the third feature mentioned above, namely, a cooperative-rationalist conception of communication[43] as a recursive reasoning process aimed at achieving communicative goals effectively.

RSA's key methodological point is a three-part communicative structure composed of (i) a literal listener, (ii) a pragmatic speaker and (iii) a pragmatic listener.[44]

The literal listener serves as the foundational layer of interpretation, determining the possibilities for mapping utterances to meanings based on syntactic and semantic constraints, without considering speaker intentions / the speaker's meaning. On this basis, the pragmatic speaker formulates an utterance by estimating its epistemic utility for the listener, i.e. by assessing the probability that the listener will successfully infer the intended meaning, given the communicative goal. This involves balancing informativeness[45] against production and interpretative effort. The pragmatic listener, in turn, estimates, based on the utterance and tassumptions about speaker rationality, the probability of an intended meaning, taking into account both the likelihood a pragmatic speaker would produce the utterance if she meant that particular meaning and the prior probability of that meaning independently of the utterance. In other words, as Judith Degen puts it, the pragmatic listener reverse engineers the speaker's most likely meaning by combining the probability of using a specific utterance with *a priori* probability of that meaning. Within the so-called uncertain RSA variant, the pragmatic speaker may also take into account a range of additional factors, including different communicative goals, background knowledge, context, the semantic contributions of smaller units to larger (utterance- or discourse-level) structures, &c.[46]

---

[41] Michael Franke and Gerhard Jäger, "Probabilistic pragmatics, or why Bayes' rule is probably important for pragmatics," *Zeitschrift für Sprachwissenschaft* 35, no. 1 (2016): 9–14.

[42] Noah D. Goodman and Michael C. Frank, "Pragmatic Language Interpretation as Probabilistic Inference," *Trends in Cognitive Sciences* 20, no. 11 (2016): 818–29; Judith Degen, "The Rational Speech Act Framework," *Annual Review of Linguistics* 9 (2023): 819–40.

[43] In this sense, although the RSA framework explicitly draws on the work of H. P. Grice, its design aligns more closely with the concepts of Relevance Theory. For instance, when Goodman and Clark define the speaker as a "utility-maximizing agent (where the effort of language production is costly, but communicating information is beneficial" (Noah D. Goodman and Michael C. Frank, "Pragmatic Language Interpretation as Probabilistic Inference," *Trends in Cognitive Sciences* 20, no. 11 (2016): 819), and more generally with the aim to "replace Grice's maxims with a single, utility-theoretic version of the cooperative principle" (Noah D. Goodman and Michael C. Frank, "Pragmatic Language Interpretation as Probabilistic Inference," *Trends in Cognitive Sciences* 20, no. 11 (2016): 821), this reflects a certain parallel to the core assumptions of Dan Sperber and Deirdre Wilson (see Dan Sperber and Deirdre Wilson, *Relevance: Communication and Cognition* (Oxford – Cambridge: Blackwell, 1995)) regarding the nature of relevance. Similar to Relevance Theory, the RSA framework effectively reduces the Gricean maxims to a single overarching universal principle.

[44] Noah D. Goodman and Michael C. Frank, "Pragmatic Language Interpretation as Probabilistic Inference," *Trends in Cognitive Sciences* 20, no. 11 (2016): 818–20; Judith Degen, "The Rational Speech Act Framework," *Annual Review of Linguistics* 9 (2023): 522.

[45] Judith Degen understands this informativeness as an attempt by the pragmatic speaker to minimize listener surprise about the meaning of the utterance. Judith Degen, "The Rational Speech Act Framework," *Annual Review of Linguistics* 9 (2023): 527.

The principal advantages of probabilistic pragmatics and RSA framework over classical pragmatic approaches lie, first, in their formal treatment of communication from the outset, and second, in their modelling of interpretation, informativeness, and listener expectations through probability and gradience rather than through categorical distinctions.[47] Beyond their testability, these approaches also align with the non-discrete, non-symbolic, and connectionist architecture of LLMs, i.e. something the traditional semiotic trichotomy is ill-equipped to accommodate. This framework captures the incremental nature of pragmatic processes, acknowledging that interpretive transitions between traditional semiotic dimensions are often gradual rather than categorical.[48] It thus allows for a view in which pragmatic effects are not confined to post-propositional inference, but manifest across all levels of meaning-making. This is becomes especially clear in phenomena such as the interpretation of referring expressions of colour terms[49], context-sensitive gradable adjectives like "strong"[50], or the non-literal interpretation of number words[51].

The de-anthropomorphising transformation of pragmatics, which is rooted in a teleological model and embodied in probabilistic approaches, signals broader departure from traditional human-centred pragmatic paradigms. Pragmatic processes, in both human communication and in LLM language processing, are not merely ex post mechanisms applied to fully formed propositions; rather they constitute a continuum of micro-adjustments that operate across all levels of the semiotic trichotomy, including the pre-, post-, and inter-propositional mechanisms previously outlined.

This shift points toward a graded ratchet than binary model of pragmatic competence. In line with Emily Sullivan's[52] functional epistemology, this approach does not regard the complexity of of pragmatic phenomena or in the "black-box" nature of LLMs as the central problem. Instead, it locates the issue in the limitations of existing theories, particularly in their failure to meaningfully specify the link between model behaviour and the relevant pragmatic phenomenon.

# 3 Substitionalism

Much (not only pragmatic) theorising about LLMs and AI tends to use these terms as umbrella terms, where arguments, whether addressing general questions of understanding or more specific issues of production and recognition of various pragmatic phenomena, often refer to this imagined abstractions rather than to the specific models. A typical example is the argument put forward by Emily Bender and Alexander Koller[53], who, in the so-called Octopus test, deny that LLMs can exhibit genuinely linguistic communicative understanding in situations that, in their view, require creative thinking, the ability to link linguistic elements with extra-linguistic referents and the capacity to grasp the speaker's communicative intentions, illustrating this claim using outputs from GPT-2. However, such a conclusion cannot be substantiated to the same extent when applied to current generations of models. The conflation of performance evaluations of a particular model with general conclusions evaluations about LLMs or AI represents a case of what I will broadly term substitutionalism, defined in a deliberately general way as the systematic yet often unreflected substitution of broader or qualitatively different elements, aspect, and properties of the communicative-pragmatic process (and of its subsequent description).

---

[46] Noah D. Goodman and Michael C. Frank, "Pragmatic Language Interpretation as Probabilistic Inference," *Trends in Cognitive Sciences* 20, no. 11 (2016): 824.
[47] Judith Degen, "The Rational Speech Act Framework," *Annual Review of Linguistics* 9 (2023): 526.
[48] Cf. Reuben Cohn-Gordon, Noah D. Goodman and Christopher Potts, "An Incremental Iterated Response Model of Pragmatics," arXiv:1810.00367 [cs.CL], 2018.
[49] Will Monroe et al., "Colors in Context: A Pragmatic Neural Model for Grounded Language Understanding," *Transactions of the Association for Computational Linguistics* 5 (2017): 325–338.
[50] Benjamin Lipkin et al., "Evaluating statistical language models as pragmatic reasoners," arXiv:2305.01020 [cs.CL], 2023.
[51] Polina Tsvilodub et al., "Non-literal Understanding of Number Words by Language Models," arXiv:2502.06204 [cs.CL], 2025.
[52] Emily Sullivan, "Understanding from Machine Learning Models," *The British Journal for the Philosophy of Science* 73, no. 1 (2022): 109–133.
[53] Emily Bender and Alexander Koller, "Climbing towards NLU: On Meaning, Form, and Understanding in the Age of Data," in *Proceedings of the 58th Annual Meeting of the Association for Computational Linguistics*, eds. Dan Jurafsky et al. (Online. Association for Computational Linguistics, 2020), 5189, 5197.

In empirically oriented pragmatic research, this form of substitutionalism occurs to a significantly lesser extent than in more theoretically inclined works. Nevertheless, questions arise concerning how researchers justify their selection of communicative subjects (especially LLMs) that serve as the primary object of investigation. In most studies, the choice (or at least its justification) of specific models tends to be abbreviated and is typically reduced either to a simple listing of those under investigation or to general selection criteria, such as diversity, popularity or the latest version. Within these pragmatically focused approaches (though similar patterns likely apply in other fields), four main parametric axes can be identified that guide the selection of models:

(1) Proprietary and open-source models. In the vast majority of cases, the principle of diversity is used to combine closed (commercial) models (especially from companies such as OpenAI, Google or Anthropic) with open (open-source) models (often those developed by companies such as Meta or Mistral). This approach aims not only to compare their performances but also to suggest that the phenomenon under investigation is general and not limited to a single model.

(2) Model size. Another application of the diversity principle lies in the inclusion of models of varying sizes (e.g. number of parameters), to account the scaling effects on the studied phenomenon and to enable comparison between smaller and larger models, as well as between earlier and more recent generations of models.

(3) Availability. In addition to positive selection criteria, negative constraints also play a role, especially regarding accessibility. A typical example is the exclusion of certain proprietary models due to lack of API access.[54]

(4) Task specificity. Finally, model choice is influenced by the nature of the phenomenon being investigated and the experimental design. While most pragmatic studies rely on domain-non-specific models considered representative of state-of-the-art capabilities, in some cases the model choice is more narrowly guided by the demands of the experiment and the output requirements. For instance, Jürgen Dietrich and André Hollstein[55] selected instruct models rather than chat models for their experiment, as the latter "tend to excuse themselves and are wordy in their outputs, which makes them suitable for use in chat applications but less favorable for data processing tasks". Conversely, R. Thomas McCoy et al.[56] found that the web-based (chat) version of GTP-4 outperformed the API version on certain tasks.

Overall, however, these parametric axes are applied in a rather unsystematic fashion. As Dojun Park et al. observe, "[d]espite the clear need for studies analysing the pragmatic competence of current LLMs, there is […] a lack of systematic evaluation across various models"[57]. One of the most characteristic symptoms of this lack of systematicity can be found in two further aspects of substitutionalism: (i) the linguistic aspect, understood as the simple uncritical substitution of specific language for language in general, and (ii) the communicative aspect, in which human subjects in communicative roles are straightforwardly substituted by LLMs.

The linguistic aspect of substitutionalism is grounded in the assumption that findings concerning one language can be extrapolated to all languages, or even to language as such. Although this issue has been acknowledged and described, most notably through a methodological imperative known as the Bender rule[58], and a partial mitigation of this limitation is provided by pragmatic benchmarks[59] that incorporate not only diverse pragmatic task types but also a range of languages, much of the existing (not only pragmatic) research still exhibits a strong English-centred bias[60].

---

[54] Dojun Park et al., "MultiPragEval: Multilingual Pragmatic Evaluation of Large Language Models," in *Proceedings of the 2nd GenBench Workshop on Generalisation (Benchmarking) in NLP*, eds. Dieuwke Hupkes et al. (Miami: Association for Computational Linguistics, 2024), 96–119.
[55] Jürgen Dietrich and André Hollstein , "Performance and Reproducibility of Large Language Models in Named Entity Recognition: Considerations for the Use in Controlled Environments," *Drug Safety* 48 (2025): 287–303.
[56] R. Thomas McCoy et al., "Embers of Autoregression: Understanding Large Language Models Through the Problem They are Trained to Solve," arXiv:2309.13638 [cs.CL], 2023, 13.
[57] Dojun Park et al., "MultiPragEval: Multilingual Pragmatic Evaluation of Large Language Models," in *Proceedings of the 2nd GenBench Workshop on Generalisation (Benchmarking) in NLP*, eds. Dieuwke Hupkes et al. (Miami: Association for Computational Linguistics, 2024), 97.

Equally pressing (and so far largely overlooked) is the communicative aspect of substitutionalism, which consists in the straightforward replacement of humans in conversational roles by LLMs. As a result, the models themselves become the focal point of inquiry, particularly regarding their performance as addressees/interpreters[61], and less frequently as speakers[62]. If we accept that interactivity between speaker and addressee is essential to pragmatic processes, then, consistent with the observation that prevailing pragmatic theories are predominantly human-centred (see Section 2), we must acknowledge that much current research has become (paradoxically) machine-centred., leaving aside the communicative behaviour of human participants in hybrid human–AI hybrid interaction.

If we take seriously the relational research agenda proposed by HMC framework (as discussed in Section 1), then just as pragmatic norms are being redefined in light of other technologies and new technological environments, especially social media[63], so too, in the case of LLMs, should we resist viewing human users merely as a reference group against which LLM performances are benchmarked.Rather, human participants should be considered as active interlocutors whose communicative behaviour is shaped and modulated by the very structure of these hybrid exchanges.

Although the incorporation of LLMs into communicative practice has led to increased communicative reflexivity among human users, manifested as increased metalinguistic and metapragmatic awareness[64], very few empirical studies have so far addressed questions, how human linguistic behaviour changes when engaging with LLMs as speakers, or what interpretative strategies humans do (or do not) apply as addressees. One such study is that of Hiromu Yakura et al.[65], which demonstrates that human spoken language is beginning to mimic LLM-like patterns. By analysing a corpus of YouTube videos from 2023 onward, the authors identified a significant increase in the frequency of expressions characteristic of ChatGPT, suggesting that people are adopting the model's linguistic features.

Despite the fact pragmatics has thus far largely ignored the linguistic and communicative behaviour of humans within AI–human exchanges, the relational and metaphysical dimensions have become particularly prominent in so-called prompt engineering[66]. This practice is not driven by efforts to approximate LLMs to human linguistic standards, but by attempts to adapt human communicative behaviour to the affordances of LLMs. Prompt engineering may be described as a set of normative guidelines for crafting prompt-oriented utterances. While a detailed analysis of this trend is beyond the scope of this section, it is worth noting that these guidelines are often based on the belief that certain forms of human linguistic behaviour can elicit more optimal or efficient responses from the model. Such instructions are not limited to structuring information, but may also target pragmatic phenomena more generally by specifying the speaker's role or other aspects of the context, or by guiding procedures in the case of more specific pragmatic categories such as politeness[67].

---

[58] Cf. Emily Bender, "The #BenderRule: On Naming the Languages We Study and Why It Matters," *The Gradient,* accessed March 3, 2025, https://thegradient.pub/the-benderrule-on-naming-the-languages-we-study-and-why-it-matters/; Emily Bender and Batya Friedman, "Data Statements for Natural Language Processing: Toward Mitigating System Bias and Enabling Better Science," *Transactions of the Association for Computational Linguistics* 6 (2018): 587–604.

[59] Cf. Dojun Park et al., "MultiPragEval: Multilingual Pragmatic Evaluation of Large Language Models," in *Proceedings of the 2nd GenBench Workshop on Generalisation (Benchmarking) in NLP*, eds. Dieuwke Hupkes et al. (Miami: Association for Computational Linguistics, 2024), 96–119.

[60] Cf. Zishan Guo et al., "Evaluating Large Language Models: A Comprehensive Survey," arXiv:2310.19736 [cs.CL], 2023; Bolei Ma et al., "Pragmatics in the Era of Large Language Models: A Survey on Datasets, Evaluation, Opportunities and Challenges". In *Proceedings of the 63rd Annual Meeting of the Association for Computational Linguistics (Volume 1: Long Papers)*, eds. Wanxiang Che et al. (Vienna: Association for Computational Linguistics, 2025), 8685.

[61] Cf. Ariane Lee, "Are GPT-3 Models Pragmatic Reasoners?" *Stanford University*, accessed March 20, 2025, https://web.stanford.edu/class/archive/cs/cs224n/cs224n.1234/final-reports/final-report-169845842.pdf

[62] Cf. Mingyue Jian and N. Siddharth, "Are LLMs good pragmatic speakers?" arXiv:2411.01562 [cs.CL], 2024.

[63] Cf. Christian R. Hoffmann and Wolfram Bublitz, eds., *Pragmatics of Social Media* (Berlin – Boston: De Gruyter Mouton, 2017); Kate Scott, *Pragmatics Online* (London – New York: Routledge, 2022).

[64] Marta Dynel, "Lessons in linguistics with ChatGPT: Metapragmatics, metacommunication, metadiscourse and metalanguage in human-AI interactions," *Language & Communication* 93 (2023): 107-124.

[65] Hiromu Yakura et al., "Empirical evidence of Large Language Model's influence on human spoken communication," arXiv:2409.01754v1 [cs.CY], 2024.

[66] Cf. e.g. Jules White et al., "A Prompt Pattern Catalog to Enhance Prompt Engineering with ChatGPT," arXiv:2302.11382 [cs.SE], 2023.

Whereas pragmatic theory tends to approach LLMs by aligning them with human communicative behaviour, prompt engineering moves in the opposite direction by guiding human users to adapt to (supposedly more) model-friendly forms of linguistic engagement. Yet neither of these approaches adequately considers the consequences of communicative substitutionalism, in which speaker and addressee roles are treated as mutually exclusive and discrete, as if either the human is the speaker and the LMM the recipient, or vice versa. This overlooks the hybrid nature of communication, in which both human and LLM can occupy both roles simultaneously (a kind of humAIn position). This hybridity allows for the emergence of communicative micro-loops, in which the user writes (or says) a prompt, the LLM generates a response, and the user then partially or fully appropriates or modifies that output with the aim of using the resulting text into other conversational interactions. Similarly, when a human acts as the recipient of model-generated reformulation, summary, or selection based on a user-provided input, the model's output may function as a form of interpretative evidence, serving as a proxy for the original text.

## 4 Context frustration

The evaluation of LLMs on selected pragmatic tasks frequently relies on multiple-choice formats, primarily due to their ease of implementation and ability to directly compare the outputs of LLMs with those of humans. While this format may be methodologically efficient and broadly interpretable, its adequacy in capturing pragmatic competence remains (to say the least) open to debate. As Shengguang Wu et al. argue, multiple-choice setups allow for cases where "a model might correctly select the option label, [but] it may still fail to respond pragmatically by itself"[68]. In contrast, open-ended evaluation methods, though significantly more challenging in terms of interpretation and consistency, may better capture the context-sensitive, inferential, and interactive nature of pragmatic understanding and processing. Yet another methodological concern arises from the frequent use of isolated prompts and artificially stylised sentences, which may distort our impression of LLMs; pragmatic abilities. If a model successfully identifies an implicature in a contrived sentence, without broader contextual anchoring, such success may reflect patterns from training data rather than genuine context-sensitive reasoning. The main challenge, then, lies not only in model capability, but in how pragmatic datasets are constructed and what assumptions underlie their design.[69]

Rather than proposing a solution to these challenges, I focus here on the conceptual question of context itself. Context is not only a central and defining concern of pragmatic theory but also critical and prominent notion in AI and LLMs technologies. However, the ways in which context is conceptualised and operationalised differ between these two domains.

The development of AI and LLMs, from basic n-gram models through attention-based transformer architectures to RAG, can be broadly viewed as a question of context, i.e. a continuous effort to address the scale, salience, and structure of preceding information. In this view, context tends to be approached technically and in the case of LLMs, we can distinguish two primary interpretations: (i) context window, i.e the maximum number of consecutive tokens the model can process at once, and (ii) training data context, i.e. the corpus from which the model derives token-level probabilities and discursive patterns. token-level probabilities are derived.

---

[67] Ziqi Yin et al., "Should We Respect LLMs? A Cross-Lingual Study on the Influence of Prompt Politeness on LLM Performance," in *Proceedings of the Second Workshop on Social Influence in Conversations (SICon 2024)*, eds. James Hale, Kushal Chawla and Muskan Garg (Miami: Association for Computational Linguistics, 2024), 9–35.

[68] Shengguang Wu et al., "Rethinking Pragmatics in Large Language Models: Towards Open-Ended Evaluation and Preference Tuning," in *Proceedings of the 2024 Conference on Empirical Methods in Natural Language Processing*, eds. Yaser Al-Onaizan, Mohit Bansal and Yun-Nung Chen (Miami: Association for Computational Linguistics, 2024), 22584.

[69] Cf. Bolei Ma et al., "Pragmatics in the Era of Large Language Models: A Survey on Datasets, Evaluation, Opportunities and Challenges". In *Proceedings of the 63rd Annual Meeting of the Association for Computational Linguistics (Volume 1: Long Papers)*, eds. Wanxiang Che et al. (Vienna: Association for Computational Linguistics, 2025), 8683–8684. Other aspects cf. e.g. Dojun Park et al., "MultiPragEval: Multilingual Pragmatic Evaluation of Large Language Models," in *Proceedings of the 2nd GenBench Workshop on Generalisation (Benchmarking) in NLP*, eds. Dieuwke Hupkes et al. (Miami: Association for Computational Linguistics, 2024), 96–119.

In pragmatics, context is typically approached as a mechanism for localising utterances in a specific situation; in this sense, context is understood, as David Lewis puts its, as "a location – time, place, and possible world – where a sentence is said"[70]. At the same time, context can also be defined, in the style of Robert Stalnaker, propositionally, as the set of background assumptions or possible worlds shared (or presumed to be shared) by participants in a communicative act.[71]

When comparing these two conceptions, one grounded in technical modelling, the other in pragmatic traditions, we encounter what may be described as a contextual paradox. On one hand, development of LLMs involves the continuous expansion of context windows and training data context, which can be interpreted as a process of massive contextualisation. If we follow a pragmatic approach to meaning as essentially dependent on contextual relations, then it seems that LLMs engage in radically non-human forms of contextual processing. Yet, this increase in data volume and window size does not resolve but rather amplifies the fundamental difference between human and machine contexts.

Following Stalnaker[72], we may say that human-LLMs interactions are contextually unstable and defective, since the speaker's and addressee's contexts (i.e., their respective sets of presuppositions or possible worlds) are not identical. Whereas Stalnaker maintains that in human-to-human communication defective contexts tend to converge towards the equilibrium of a non-defective context, in human-LLMs interactions such equilibrium is in principle unattainable. This is because the sets of contexts/presuppositions/possible worlds available to LLMs are incommensurable with those accessible to human interlocutors.

This situation is further complicated by the role of alignment processes, such as instruct tuning and reinforcement learning from human feedback (RLHF), which significantly shape the pragmatic behaviour of LLMs (as demonstrated by pragmatic research itself[73]). Models may refuse to generate certain outputs not due to contextual under-specification in the technical sense, but because such responses are blocked by additional training. As Marta Andersson and Dan McIntyre[74] show, ChatGPT-3.5 is relatively successful in identifying impoliteness in conventionalised contexts (whether literal or inferential), but performs less consistently when the offensive content is indirect, inferential or situated. The model's increased sensitivity to potential harm contrasts with human flexibility and suggests that certain pragmatic mechanisms, especially those based on implicature, are unevenly represented in model behaviour, even when additional context is provided.

Prompt engineering handbooks frequently recommend that users provide clear contextual anchoring within their prompts, most often by specifying the Lewisian localisation parameters, such as assigning the model a specific role (e.g., teacher, critic &c.). This also implicitly defines the user's role, albeit in less determinate terms (for example, as student rather than fellow teacher). While these strategies aim to stabilise context (in Stalnaker's sense) and reduce contextual asymmetry, they often carry the risk of reinforcing substitutionalist assumptions, i.e. replacing complex, dynamic human roles with simplified, static positions for the sake of clarity.

Even though some studies[75] have pointed to the correlation between model size (see also Section 3) and performance on pragmatic tasks, the deeper issue lies in the limits of technical contextualisation itself. As I have argued in Section 1, LLMs challenge the discrete semiotic segmentation of meaning, including its pragmatic component. John Searle's[76] claim, that context cannot be represented by a finite set of propositions and that background assumptions are neither expressible nor fully formalizable, implies that no increase in token limits or training data scale can ever fully exhaust or capture context. From a pragmatic standpoint, this constitutes a more serious critique of AI and LLMs comprehension than Searle's well-known Chinese Room thought experiment.

---

[70] David Lewis, "Index, Context, and Content," in *Philosophy and Grammar,* eds. Stig Kanger and Sven Öhman (Dordrecht – Boston – London: D. Reidel, 1980): 79.

[71] Robert C. Stalnaker, *Context and Content* (Oxford: Oxford University Press, 1999): 84.

[72] Robert C. Stalnaker, *Context and Content* (Oxford: Oxford University Press, 1999): 85.

[73] Cf. Laura Ruis et al., "The Goldilocks of Pragmatic Understanding: Fine-Tuning Strategy Matters for Implicature Resolution by LLMs," arXiv:2210.14986 [cs.CL], 2023;

[74] Marta Andersson and Dan McIntyre, "Can ChatGPT recognize impoliteness? An exploratory study of the pragmatic awareness of a large language model," *Journal of Pragmatics* 239 (2025): 16-36.

[75] Polina Tsvilodub et al, "Experimental Pragmatics with Machines: Testing LLM Predictions for the Inferences of Plain and Embedded Disjunctions," arXiv:2405.05776 [cs.CL], 2024.

[76] John Searle, *Expression and Meaning* (Cambridge: Cambridge University Press, 1979): 125.

The second component of the contextual paradox stems from the so-called context collapse[77]. Focusing on human-to-human communication on social media, this term attempts to describe a situation in which "social media collapse diverse social contexts into one, making it difficult for people to engage in the complex negotiations needed to vary identity presentation, manage impressions, and save face"[78]. As speakers are pulled into hybrid, heterogeneous communicative environments by virtue of the technological infrastructures they use, context becomes fractured, defined less by shared assumptions and more by competing interpretive frameworks.[79] This heterogeneity is not limited to multilingualism or code-switching, but also includes divergent cultural and normative expectations among audiences.

In communicative situations involving LLMs, context collapse manifests not only through the tension created by the incredibly diverse types of audiences who are (or may be) recipients of LLM outputs, but also through the fluidity of speaker roles that LLMs can, in principle, occupy. Human–LLMs communication is therefore hybrid not only because it blurs the boundaries between human and machine contributions, but also because it reshapes the structural topological structure of communication itself. Human-LLMs interaction is not limited to traditional many (people) – to – one (model) or one (model) – to – many (people) configurations, but increasingly resemble new type of many-to-many arrangements. These are characterised by chained exchanges "prompt – reply – re-prompt" and by communicative micro-loops (see Section 3), where interpretation and generation are distributed across recursive iterations of co-production.

The simultaneous processes of massive contextualisation and massive context collapse are what I propose to call context frustration. This term aims to capture the experiential dissonance (or emotional frustration) felt by human users who engage in exchanges where contextual alignment repeatedly fails, where the context is defective (in the Lewisian sense) and not mutually shared, and where attempts to clarify, specify, or stabilise context (by filling the context window) are undone by topological complexity and structural limitations of the exchange. Context frustration also captures the cognitive tension arising from the failure to localise contextual (in the Lewisian sense) aspects of utterance and discourse. This tension is not only epistemic (see Section 2), but also structural. It emerges from the architecture of the conversation itself., where both human and models participate in co-creating meaning and conditions of meaningfulness, yet remain partially dislocated in terms of shared reference, inference, and communicative expectations.

# 5 Conclusion

In this paper, I have sought to argue that pragmatics should not be understood merely as the third and hierarchically distinct dimension of meaningfulness, but rather as a dynamic interface through which language becomes a socially anchored instrument of action. However, with the entry of LLMs into communicative practices, this understanding must be further differentiated and methodologically reconsidered.

In the first section, I attempted to show that the semiotic trichotomy and the traditional position of pragmatics within it begins to dissolve in the era of connectionist LLM architectures. It becomes necessary, therefore, to explore alternative frameworks for studying meaningfulness, such as the communicatively oriented HMC framework, which (with its functional, relational, and metaphysical aspects) offers a potentially more adequate methodological structure than the concept of a pragmatic "wastebasket" that allegedly concludes the meaning-making process.

In the second section, I aimed to highlight the tension between human-centred pragmatic theories and the machine-centred object of their inquiry. Although traditional pragmatic frameworks (particularly of the Gricean type) continue to structure both theoretical and empirical approaches to LLMs, they remain bound to the logic of the semiotic hierarchy, treating pragmatic mechanisms primarily as post-propositional, and relying on specifically human communicative assumptions that differ from the predictive nature of LLMs. From this perspective, probabilistic pragmatics, and especially the Rational Speech Act framework, appears to provide a teleological framework more attuned to the character of LLMs, which are not "truth-evaluating" machines but rather machines designed to optimise distributional probabilities within the limits of available computational resources.

---

[77] Alice E. Marwick and danah boyd, "I tweet honestly, I tweet passionately: Twitter users, context collapse, and the imagined audience," *New Media & Society* Volume 13, no 1 (2011): 114–133; danah boyd, "Social Network Sites as Networked Publics. Affordances, Dynamics, and Implications," in *A Networked Self. Identity, Community, and Culture on Social Networks Sites*, ed. Zizi Papacharissi (New York – London: Routledge, 2011), 50–51.

[78] Alice E. Marwick and danah boyd, "I tweet honestly, I tweet passionately: Twitter users, context collapse, and the imagined audience," *New Media & Society* Volume 13, no 1 (2011): 123.

[79] Cf. Jannis Androutsopoulos, "Languaging when contexts collapse: Audience design in social networking," *Discourse, Context and Media* 4–5 (2014): 62–73.

In the third section, I attempted to delineate the problem of substitutionalism and its three forms: (i) generalising, where the performance of specific model is interpreted as representative of AI or LLMs in general; (ii) linguistic, where a particular language (English) is substituted for language as such; and (iii) communicative, where, mainly in in experimental designs, human communicative participants are straightforwardly substituted by LLMs. This leads, on the one hand, to LLM performance being assessed using anthropomorphic criteria, while on the other hand systematically overlooking those communicative subjects (i.e. humans) for whom such theories might, in fact, be more appropriately applied.

In the final, fourth part, I tried to show that the massive expansion of the context window and the growth of training data volume go hand in hand with a massive collapse of context. I refer to this situation as "context frustration". It manifests not only as an asymmetry of presuppositions between human users and the model, but also as pressure on human users to engage in specific linguistic behaviours aimed at explicitly fixing selected aspects of context. In doing so, they co-create not only the communicative and pragmatic conditions for the models with which they interact, but also for their own linguistic performances, which they may subsequently carry over into non-LLM conversations.

**Acknowledgment**
This work has been funded by a grant from the Programme Johannes Amos Comenius under the Ministry of Education, Youth and Sports of the Czech Republic, CZ.02.01.01/00/23_025/0008711.

# References

Andersson, Marta and Dan McIntyre. "Can ChatGPT recognize impoliteness? An exploratory study of the pragmatic awareness of a large language model." *Journal of Pragmatics* 239 (2025): 16-36. https://doi.org/10.1016/j.pragma.2025.02.001

Androutsopoulos, Jannis. "Languaging when contexts collapse: Audience design in social networking." *Discourse, Context and Media* 4–5 (2014): 62–73. https://doi.org/10.1016/j.dcm.2014.08.006

Baroni, Marco. "On the proper role of linguistically-oriented deep net analysis in linguistic theorizing." In *Algebraic systems and the representation of linguistic knowledge*, edited by Shalom Lappin, 5–22. Abingdon-on-Thames: Taylor and Francis, 2022.

Basile, Pierpaolo et al. "Exploring the Word Sense Disambiguation Capabilities of Large Language Models." arXiv:2503.08662 [cs.CL], 2025. https://doi.org/10.48550/arXiv.2503.08662

Bender, Emily and Batya Friedman. "Data Statements for Natural Language Processing: Toward Mitigating System Bias and Enabling Better Science." *Transactions of the Association for Computational Linguistics* 6 (2018): 587–604. https://doi.org/10.1162/tacl_a_00041

Bender, Emily. "The #BenderRule: On Naming the Languages We Study and Why It Matters." *The Gradient*. Article published September 19, 2019. https://thegradient.pub/the-benderrule-on-naming-the-languages-we-study-and-why-it-matters/.

Bender, Emily. "Thought experiment in the National Library of Thailand." *Medium*. Article published May 25, 2023. https://medium.com/@emilymenonbender/thought-experiment-in-the-national-library-of-thailand-f2bf761a8a83.

Bender, Emily and Alexander Koller. "Climbing towards NLU: On Meaning, Form, and Understanding in the Age of Data." In *Proceedings of the 58th Annual Meeting of the Association for Computational Linguistics*, edited by Dan Jurafsky et al., 5185–5198. Online. Association for Computational Linguistics, 2020. https://doi.org/10.18653/v1/2020.acl-main.463

Biletzki, Anat. "Is there a history of pragmatics?" *Journal of Pragmatics* 25, no. 4 (1996): 455-470. https://doi.org/10.1016/0378-2166(95)00019-4

Bojić, Ljubiša, Predrag Kovačević and Milan Čabarkapa. "Does GPT-4 surpass human performance in linguistic pragmatics?" *Humanities and Social Sciences Communications* 12, 794 (2025). https://doi.org/10.1057/s41599-025-04912-x

Bolei Ma et al. "Pragmatics in the Era of Large Language Models: A Survey on Datasets, Evaluation, Opportunities and Challenges". In *Proceedings of the 63rd Annual Meeting of the Association for Computational Linguistics (Volume 1: Long Papers)*, edited by Wanxiang Che et al., 8679–8696. Vienna: Association for Computational Linguistics, 2025. https://doi.org/10.18653/v1/2025.acl-long.425

Borg, Emma. *Minimal Semantics.* Oxford: Oxford University Press, 2004.

Borg, Emma. *Pursuing Meaning.* Oxford: Oxford University Press, 2012.

Borg, Emma. "LLMs, Turing tests and Chinese rooms: the prospects for meaning in large language models," *Inquiry* (2025). https://doi.org/10.1080/0020174X.2024.2446241

boyd, danah. "Social Network Sites as Networked Publics. Affordances, Dynamics, and Implications." In *A Networked Self. Identity, Community, and Culture on Social Networks Sites*, edited by Zizi Papacharissi, 39–58. New York – London: Routledge, 2011.

Cappelen, Herman and Ernie Lepore. *Insensitive Semantics.* Malden – Oxford – Carlton: Blackwell, 2005.

Cappelen, Herman and Josh Dever. *Bad Language.* Oxford: Oxford University Press, 2019.

Cappelen, Herman and Josh Dever. "AI with Alien Content and Alien Metasemantics." In *The Oxford Handbook of Applied Philosophy of Language*, edited by Ernie Lepore and Luvell Anderson, 573–593. Oxford: Oxford University Press, 2024.

Carenini, Gaia et al. "Large Language Models Behave (Almost) As Rational Speech Actors: Insights From Metaphor Understanding." In *Open Review / Information-Theoretic Principles in Cognitive Systems Workshop at 37th Conference on Neural Information Processing Systems (NeurIPS 2023)*, Article published October 27, 2023. https://openreview.net/pdf?id=SosbRhZLBV.

Carnap, Rudolf. *Introduction to Semantics*. Cambridge: Harvard University Press, 1948.

Carston, Robyn. *Thoughts and Utterances.* Oxford – Berlin: Blackwell, 2002.

Chen, Xi Jun Li and Yuting Ye. "A feasibility study for the application of AI-generated conversations in pragmatic analysis." *Journal of Pragmatics* 223 (2024): 14-30. https://doi.org/10.1016/j.pragma.2024.01.003

Cohn-Gordon, Reuben; Noah D. Goodman and Christopher Potts. "An Incremental Iterated Response Model of Pragmatics." arXiv:1810.00367 [cs.CL]. 2018. https://doi.org/10.48550/arXiv.1810.00367

Cong, Yan. "Manner implicatures in large language models." *Sci Rep* 14, 29113 (2024). https://doi.org/10.1038/s41598-024-80571-3

Degen, Judith. "The Rational Speech Act Framework." *Annual Review of Linguistics* 9 (2023): 519–40. https://doi.org/10.1146/annurev-linguistics-031220-010811

Dentella, Vittoria et al. "Testing AI on language comprehension tasks reveals insensitivity to underlying meaning." *Scientific Reports* 14, 28083 (2024). https://doi.org/10.1038/s41598-024-79531-8

Devitt, Michael. *Overlooking Conventions.* Cham: Springer, 2021.

Dietrich, Jürgen and André Hollstein. "Performance and Reproducibility of Large Language Models in Named Entity Recognition: Considerations for the Use in Controlled Environments," *Drug Safety* 48 (2025): 287–303. https://doi.org/10.1007/s40264-024-01499-1

Dijk, Bram van et al. "Large Language Models: The Need for Nuance in Current Debates and a Pragmatic Perspective on Understanding." In *Proceedings of the 2023 Conference on Empirical Methods in Natural Language Processing*, edited by Houda Bouamor, Juan Pino, Kalika Bali, 12641–12654. Singapore: Association for Computational Linguistics, 2023. https://doi.org/10.18653/v1/2023.emnlp-main.779

Dynel, Marta. "Lessons in linguistics with ChatGPT: Metapragmatics, metacommunication, metadiscourse and metalanguage in human-AI interactions." *Language & Communication* 93 (2023): 107-124. https://doi.org/10.1016/j.langcom.2023.09.002

Erk, Katrin. "The Probabilistic Turn in Semantics and Pragmatics." *Annual Review of Linguistics* 8 (2022):101–21. https://doi.org/10.1146/annurev-linguistics-031120-015515

Etzrodt, Katrin et al. "Human-machine-communication: introduction to the special issue." *Publizistik* 67 (2022): 439–448. https://doi.org/10.1007/s11616-022-00754-8

Flamino, James et al. "Testing the limits of large language models in debating humans." *Scientific Reports* 15, 13852 (2025). https://doi.org/10.1038/s41598-025-98378-1

Floridi, Luciano. "AI as Agency Without Intelligence: on ChatGPT, Large Language Models, and Other Generative Models." *Philosophy and Technology* 36, 15 (2023). https://doi.org/10.1007/s13347-023-00621-y

Franke, Michael and Gerhard Jäger. "Probabilistic pragmatics, or why Bayes' rule is probably important for pragmatics." *Zeitschrift für Sprachwissenschaft* 35, no. 1 (2016): 3–44. https://doi.org/10.1515/zfs-2016-0002

Gazdar, Gerald. *Pragmatics*. New York – London: Academic Press, 1979.

Grice, Paul. *Studies in the Way of Words.* Cambridge – London: Harvard University Press, 1989.

Goodman, Noah D. and Michael C. Frank. "Pragmatic Language Interpretation as Probabilistic Inference." *Trends in Cognitive Sciences* 20, no. 11 (2016): 818–29. https://doi.org/10.1016/j.tics.2016.08.005

Gubelmann, Reto. "Large Language Models, Agency, and Why Speech Acts are Beyond Them (For Now) – A Kantian-Cum-Pragmatist Case." *Philosophy & Technology* 37, 32 (2024). https://doi.org/10.1007/s13347-024-00696-1

Gubelmann, Reto. "Pragmatic Norms Are All You Need – Why The Symbol Grounding Problem Does Not Apply to LLMs." In *Proceedings of the 2024 Conference on Empirical Methods in Natural Language Processing*, 11663–11678. Miami: Association for Computational Linguistics, 2024. https://doi.org/10.18653/v1/2024.emnlp-main.651

Gunkel, David J. *AI for Communication.* Boca Raton: CRC Press, 2025.

Guo, Zishan et al. "Evaluating Large Language Models: A Comprehensive Survey." arXiv:2310.19736 [cs.CL], 2023. https://doi.org/10.48550/arXiv.2310.19736

Guzman, Andrea L. ed. *Human-Machine Communication. Rethinking Communication, Technology, and Ourselves.* New York: Peter Lang, 2018.

Guzman, Andrea L. and Seth C. Lewis. "Artificial intelligence and communication: A Human–Machine Communication research agenda." *New Media & Society* 22, no. 1 (2019): 70-86. https://doi.org/10.1177/1461444819858691

Harnad, Steven. "The Symbol Grounding Problem." *Physica* D 42 (1990): 335–346.

Havlík, Vladimír, "Meaning and understanding in large language models." *Synthese* 205, no. 9 (2025). https://doi.org/10.1007/s11229-024-04878-4

Hoffmann, Christian R. and Wolfram Bublitz, eds. *Pragmatics of Social Media*. Berlin – Boston: De Gruyter Mouton, 2017.

Jian, Mingyue and N. Siddharth. "Are LLMs good pragmatic speakers?" arXiv:2411.01562 [cs.CL]. 2024. https://doi.org/10.48550/arXiv.2411.01562

Kallini, Julie et al. "Mission: Impossible Language Models." In *Proceedings of the 62nd Annual Meeting of the Association for Computational Linguistics (Volume 1: Long Papers)*, edited by Lun-Wei Ku, Andre Martins and Vivek Srikumar, 14691–14714. Bangkok: Association for Computational Linguistics, 2024. https://doi.org/10.18653/v1/2024.acl-long.787

Kamath, Gaurav et al. "Scope Ambiguities in Large Language Models." *Transactions of the Association for Computational Linguistics* 12 (2024): 738–754. https://doi.org/10.1162/tacl_a_00670

Kaplan, David. "Afterthoughts." In *Themes from Kaplan*, edited by Joseph Almog, John Perry and Howard Wettstein. 565–614. New York – Oxford: Oxford University Press, 1989.

Keiser, Jessica. *Non-Ideal Foundations of Language.* New York: Routledge, 2023.

Kim, Najoung and Sebastian Schuster. "Entity Tracking in Language Models." In *Proceedings of the 61st Annual Meeting of the Association for Computational Linguistics (Volume 1: Long Papers)*, edited by Anna Rogers, Jordan Boyd-Graber and Naoaki Okazaki, 3835–3855. Toronto: Association for Computational Linguistics, 2023. https://doi.org/10.18653/v1/2023.acl-long.213

Koyama, Wataru. "The rise of pragmatics: a historiographic overview." in *Foundations of Pragmatics*, edited by Wolfram Bublitz and Neal R. Norrick, 139–166. Berlin – Boston: De Gruyter Mouton, 2011.

Lipkin, Benjamin et al. "Evaluating statistical language models as pragmatic reasoners." arXiv:2305.01020 [cs.CL]. 2023. https://doi.org/10.48550/arXiv.2305.01020

Lee, Ariane. "Are GPT-3 Models Pragmatic Reasoners?" *Stanford University*. Article accessed March 20, 2025. https://web.stanford.edu/class/archive/cs/cs224n/cs224n.1234/final-reports/final-report-169845842.pdf

Lewis, David. "Index, Context, and Content." In *Philosophy and Grammar*, eds. Stig Kanger and Sven Öhman, 79–100. Dordrecht – Boston – London: D. Reidel, 1980.

Mao, Rui et al. "A survey on pragmatic processing techniques." *Information Fusion* 114, 102712 (2025). https://doi.org/10.1016/j.inffus.2024.102712

Marwick, Alice E. and danah boyd. "I tweet honestly, I tweet passionately: Twitter users, context collapse, and the imagined audience." *New Media & Society* 13, no. (2011): 114–133. https://doi.org/10.1177/1461444810365313

McCoy, R. Thomas et al. "Embers of Autoregression: Understanding Large Language Models Through the Problem They are Trained to Solve." arXiv:2309.13638 [cs.CL]. 2023. https://doi.org/10.48550/arXiv.2309.13638

Min, Sewon et al. "AmbigQA: Answering Ambiguous Open-domain Questions." In *Proceedings of the 2020 Conference on Empirical Methods in Natural Language Processing (EMNLP)*, edited by Bonnie Webber et al., 5783–5797. Online: Association for Computational Linguistics, 2020. https://doi.org/10.18653/v1/2020.emnlp-main.466

Mollo, Dimitri Coelho and Raphaël Millière. "The Vector Grounding Problem." arXiv:2304.01481 [cs.CL]. 2023. https://doi.org/10.48550/arXiv.2304.01481

Monroe, Will et al. "Colors in Context: A Pragmatic Neural Model for Grounded Language Understanding." *Transactions of the Association for Computational Linguistics* 5 (2017): 325–338. https://doi.org/10.1162/tacl_a_00064

Moro, Andrea; Greco, Matteo and Stefano F. Cappa. "Large languages, impossible languages and human brains." *Cortex* 167 (2023): 82–85. https://doi.org/10.1016/j.cortex.2023.07.003

Morris, Charles W. *Foundations of the Theory of Signs*. Chicago: University of Chicago Press, 1938.

Nizamani, Rashid, Sebastian Schuster and Vera Demberg. "SIGA: A Naturalistic NLI Dataset of English Scalar Implicatures with Gradable Adjectives." In *Proceedings of the 2024 Joint International Conference on Computational Linguistics, Language Resources and Evaluation (LREC-COLING 2024)*, edited by Nicoletta Calzolari et al., 14784–14795. Torino: ELRA and ICC, 2024.

Noveck, Ira A. and Anne Reboul. "Experimental Pragmatics: a Gricean turn in the study of language." *Trends in Cognitive Science* 12, no. 11 (2008): 425–431. https://doi.org/10.1016/j.tics.2008.07.009

Noveck, Ira A. and Dan Sperber, eds. *Experimental Pragmatics.* Basingstoke – New York: Palgrave, 2004.

Park, Dojun et al. "MultiPragEval: Multilingual Pragmatic Evaluation of Large Language Models." In *Proceedings of the 2nd GenBench Workshop on Generalisation (Benchmarking) in NLP*, edited by Dieuwke Hupkes et al., 96–119. Miami: Association for Computational Linguistics, 2024. https://doi.org/10.18653/v1/2024.genbench-1.7

Peregrin, Jaroslav. "Do Computers "Have Syntax, But No Semantics"?" *Minds and Machines* 31, no. 2 (2021): 305–321. https://doi.org/10.1007/s11023-021-09564-9

Piantadosi, Steven T. and Felix Hill. "Meaning without reference in large language models," arXiv:2208.02957 [cs.CL]. 2022. https://doi.org/10.48550/arXiv.2208.02957

Qiu, Zhuang; Xufeng Duan and Zhenguang Cai. "Does ChatGPT Resemble Humans in Processing Implicatures?" In *Proceedings of the 4th Natural Logic Meets Machine Learning Workshop (NALOMA23)*, edited by Stergios Chatzikyriakidis and Valeria de Paiva, 25–34. Nancy: Association for Computational Linguistics, 2023.

Radivojevic, Kristina, Nicholas Clark and Paul Brenner. "LLMs Among Us: Generative AI Participating in Digital Discourse." arXiv:2402.07940 [cs.HC]. 2024. https://doi.org/10.48550/arXiv.2402.07940

Recanati, Francois. *Literal Meaning.* New York: Cambridge University Press, 2004.

Recanati, Francois. *Truth-Conditional Pragmatics.* Oxford: Clarendon Press, 2010.

Rosen, Zachary P and Rick Dale. "LLMs Don't "Do Things with Words" but Their Lack of Illocution Can Inform the Study of Human Discourse." *Proceedings of the Annual Meeting of the Cognitive Science Society* 46 (2024). https://escholarship.org/uc/item/25k7z0mz

Ruis, Laura et al. "The Goldilocks of Pragmatic Understanding: Fine-Tuning Strategy Matters for Implicature Resolution by LLMs." arXiv:2210.14986 [cs.CL]. 2023. https://doi.org/10.48550/arXiv.2210.14986

Scott, Kate. *Pragmatics Online.* London – New York: Routledge, 2022.

Searle, John. *Expression and Meaning.* Cambridge: Cambridge University Press, 1979.

Searle, John. "Minds, brains, and programs." *Behavioral and Brain Sciences* 3, no. 3 (1980): 417–457. https://doi.org/10.1017/S0140525X00005756

Shisen, Yue et al. "Do Large Language Models Understand Conversational Implicature- A case study with a Chinese sitcom." In *Proceedings of the 23rd Chinese National Conference on Computational Linguistics (Volume 1: Main Conference)*, edited by Maosong Sun et al., 1270–1285. Taiyuan: Chinese Information Processing Society of China, 2024.

Sperber, Dan and Deirdre Wilson. *Relevance. Communication and Cognition.* Oxford – Cambridge: Blackwell, 1995.

Sravanthi, Settaluri et al. "PUB: A Pragmatics Understanding Benchmark for Assessing LLMs' Pragmatics Capabilities." In *Findings of the Association for Computational Linguistics: ACL 2024*, edited by Lun-Wei Ku, Andre Martins and Vivek Srikumar, 12075–12097. Bangkok: Association for Computational Linguistics, 2024. https://doi.org/10.18653/v1/2024.findings-acl.719

Stalnaker Robert C. *Context and Content.* Oxford: Oxford University Press, 1999.

Sullivan, Emily. "Understanding from Machine Learning Models" *The British Journal for the Philosophy of Science* 73, no. 1 (2022): 109–133. https://doi.org/10.1093/bjps/axz035

Sundar, S Shyam and Eun-Ju Lee. "Rethinking Communication in the Era of Artificial Intelligence." *Human Communication Research* 48, no. 3 (2022): 379–385. https://doi.org/10.1093/hcr/hqac014

Tsvilodub, Polina et al. "Experimental Pragmatics with Machines: Testing LLM Predictions for the Inferences of Plain and Embedded Disjunctions." arXiv:2405.05776 [cs.CL]. 2024. https://doi.org/10.48550/arXiv.2405.05776

Tsvilodub, Polina et al., "Non-literal Understanding of Number Words by Language Models," arXiv:2502.06204 [cs.CL]. 2025. https://doi.org/10.48550/arXiv.2502.06204

Turing, Alan. "Computing Machinery and Intelligence." *Mind* 59, no. 236 (1950): 433–460. https://doi.org/10.1093/mind/LIX.236.433

White, Jules et al. "A Prompt Pattern Catalog to Enhance Prompt Engineering with ChatGPT." arXiv:2302.11382 [cs.SE]. 2023. https://doi.org/10.48550/arXiv.2302.11382

Williams, Iwan and Tim Bayne. "Chatting with bots: AI, speech acts, and the edge of assertion," *Inquiry* (2024): 1–24. https://doi.org/10.1080/0020174X.2024.2434874

Wu, Shengguang et al. "Rethinking Pragmatics in Large Language Models: Towards Open-Ended Evaluation and Preference Tuning" In *Proceedings of the 2024 Conference on Empirical Methods in Natural Language Processing*, edited by Yaser Al-Onaizan, Mohit Bansal and Yun-Nung Chen, 22583–22599. Miami: Association for Computational Linguistics, 2024. https://doi.org/10.18653/v1/2024.emnlp-main.1258

Yakura Hiromu et al. "Empirical evidence of Large Language Model's influence on human spoken communication." arXiv:2409.01754v1 [cs.CY]. 2024. https://doi.org/10.48550/arXiv.2409.01754

Yin, Ziqi et al. "Should We Respect LLMs? A Cross-Lingual Study on the Influence of Prompt Politeness on LLM Performance," in *Proceedings of the Second Workshop on Social Influence in Conversations (SICon 2024)*, edited by James Hale, Kushal Chawla and Muskan Garg, 9–35. Miami: Association for Computational Linguistics, 2024. https://doi.org/10.18653/v1/2024.sicon-1.2

Yu, Danni et al. "Assessing the potential of LLM-assisted annotation for corpus-based pragmatics and discourse analysis: The case of apology." *International Journal of Corpus Linguistics* 29, no 4 (2024): 534–561. https://doi.org/10.1075/ijcl.23087.yu

Zhang, Min et al. "Don't Go To Extremes: Revealing the Excessive Sensitivity and Calibration Limitations of LLMs in Implicit Hate Speech Detection." In *Proceedings of the 62nd Annual Meeting of the Association for Computational Linguistics (Volume 1: Long Papers)*, edited by Lun-Wei Ku, Andre Martins, Vivek Srikumar, 12073–12086. Bangkok: Association for Computational Linguistics, 2024. https://doi.org/10.18653/v1/2024.acl-long.652